\pdfoutput=1

\documentclass[11pt]{article}

\usepackage{acl}

\usepackage{times}
\usepackage{latexsym}
\usepackage{todonotes}
\usepackage[T1]{fontenc}

\usepackage[utf8]{inputenc}

\usepackage{microtype}

\usepackage{subcaption}
\usepackage{amsmath}
\usepackage{graphicx}

\title{SubER: A Metric for Automatic Evaluation of Subtitle Quality}

\author{Patrick Wilken \\
  AppTek  \\
  Aachen, Germany \\
  \texttt{\small pwilken@apptek.com} \\\And
  Panayota Georgakopoulou \\
  Athena Consultancy \\
  Athens, Greece \\
  \texttt{\small yota@athenaconsultancy.eu} \\\And  
  Evgeny Matusov \\
  AppTek  \\
  Aachen, Germany \\
  \texttt{\small ematusov@apptek.com} \\}

\begin{document}
\maketitle
\begin{abstract}
This paper addresses the problem of evaluating 
the quality of automatically
generated subtitles, which includes not only
the quality of the machine-transcribed or translated speech,
but also the quality of line segmentation and subtitle timing. 
We propose SubER - a single novel metric based on edit distance with shifts that takes all of these subtitle properties into account. We compare it to existing metrics for evaluating transcription, translation, and subtitle quality. A careful human evaluation in a post-editing scenario shows that the new metric has a high correlation with the post-editing effort and direct human assessment scores, outperforming baseline metrics considering only the subtitle text, such as WER and BLEU, and existing methods to integrate segmentation and timing features.
\end{abstract}

\section{Introduction}

The use of automatically created subtitles has become popular due to improved speech recognition (ASR) and machine translation (MT) quality in recent years. Most notably, they are used on the web to make content available to a broad audience in a cost-efficient and scalable way. They also gain attraction in the media industry, where they can be an aid to professional subtitlers and lead to increased productivity.

In this work, we address the problem of measuring the quality of such automatic subtitling systems. We argue that existing metrics which compare the plain text output of an ASR or MT system to a reference text are not sufficient to reflect the particularities of the subtitling task. We consider two use cases: 1) running speech recognition on the audio track of a video to create subtitles in the original language; 2) translating existing subtitle files with an MT system. For the first case, the word error rate (WER) of the ASR system is a natural choice for quality control. For MT there exist a wider range of automatic metrics such as BLEU \citep{papineni-etal-2002-bleu}, TER \citep{snover-etal-2006-study}, chrF \citep{popovic-2015-chrf} and, more recently, learned metrics like BertScore \citep{zhang2019bertscore} and COMET \citep{rei-etal-2020-comet}.

These existing metrics are suited to measure the quality of ASR and MT in terms of recognized or translated content only. However, subtitles are defined by more than just their textual content: they include timing information, as well as formatting with possible line breaks within a sentence in syntactically and semantically proper positions.
Figure \ref{fig:srt_example} shows examples of subtitle files in the common SubRip text (SRT) format. Evidently, it differs from plain text, in particular:
\begin{itemize}
    \itemsep0em 
    \item The text is segmented into blocks. These blocks are distinct from sentences. A sentence can span several blocks, a block can contain multiple sentences.
    \item A block may be further split into lines.
    \item Start and end times define when text is displayed.
\end{itemize}

\begin{figure*}[h]
    \begin{subfigure}[t]{.5\textwidth}
    \small
    \tt
    \fbox{\parbox{0.93\linewidth}{
    694 \\
00:50:45,500 --> 00:50:47,666 \\
For the brandy and champagne  \\
you bought me. \\
\\
695 \\
00:50:47,750 --> 00:50:51,375 \\
As I remember, it was the booze that \\
put you to sleep a little prematurely.\\
\\
696 \\
00:50:52,208 --> 00:50:54,291 \\
Ladies and gentlemen, \\
\\
697 \\
00:50:54,916 --> 00:50:57,291 \\
the dance is about to begin. \\
    }}
    \end{subfigure}
        \begin{subfigure}[t]{.5\textwidth}
    \small
    \tt
    \fbox{\parbox{0.93\linewidth}{
634 \\
00:50:44,960 --> 00:50:47,680 \\
For the champagne \\
and brandy you bought me. \\

635 \\
00:50:47,760 --> 00:50:51,200 \\
As I recall, the booze put you \\
to sleep a little prematurely. \\

636 \\
00:50:52,200 --> 00:50:57,120 \\
Ladies and gentlemen, \\
the dance is about to begin. \\
\\
\\
\\
    }}
    \end{subfigure}
    \caption{Two examples of subtitles in SRT format for the same video excerpt. Note the different line and block segmentation. Also note that subtitles on the right have been condensed for improved readability.}
    \label{fig:srt_example}
    \vspace{-0.2cm}
\end{figure*}

All of these additional characteristics are crucial for the viewers' comprehension of the content. Professional subtitlers check and possibly improve them as part of the machine-assisted process of subtitle creation.

To assess the quality of automatically created subtitle files, it is beneficial to have a \textit{single} metric that evaluates the ASR/MT quality and the quality of the characteristics listed above.

The main contributions of this work are:
\begin{enumerate}
    \itemsep0em 
    \item A novel segmentation- and timing-aware quality metric designed for the task of automatic subtitling.
    \item A human evaluation that analyzes
    how well the proposed metric correlates with human judgements of subtitle quality, measured in post-editing effort as well as direct assessment scores.
    \item The publication of a scoring tool to calculate the proposed metric as well as many baseline metrics, directly operating on subtitle files:\\ {\small \url{https://github.com/apptek/SubER}}
\end{enumerate}

\section{Subtitle Quality Assessment in the Media Industry}
Related to this work are subtitling quality metrics used in the media industry.
The most widely used ones to date are NER~\cite{romeroperez2015} and NTR~\cite{romero2017}  for live subtitle quality, the former addressing intralingual subtitles or captions and the latter interlingual ones.

Offline interlingual subtitles have traditionally been assessed on the basis of internal quality guidelines and error typologies produced by media localization companies. To address this gap, the FAR model~\cite{pedersen2017} was developed and there have also been attempts to implement a version of MQM\footnote{Multidimensional Quality Metrics (MQM) Definition http://www.qt21.eu/mqm-definition/definition-2015-12-30.html}.

None of the above metrics, however, are automatic ones. They require manual evaluation by an expert to categorize errors and assign appropriate penalties depending on their severity. This makes their use costly and time-consuming. In this work we therefore address automatic quality assessment of subtitle files by comparing them to a professionally created reference.

\section{Automatic Metrics for Subtitling}
\subsection{Baseline Approaches}
\label{subsec:base_metrics}
When subtitling in the original language of a video, the baseline quality measurement is to calculate word error rate (WER) against a reference transcription. Traditionally, WER is computed on lower-cased words and without punctuation. We show results for a cased and punctuated variant as well, as those are important aspects of subtitle quality. Because of the efficiency of the Levenshtein algorithm, WER calculation can be done on the whole file without splitting it into segments.

For translation, automatic metrics are usually computed on sentence level. \citet{karakanta-etal-2020-42} and other related work assumes hypothesis-reference sentence pairs to be given for subtitle scoring. However, in the most general case we only have access to the reference subtitle file and the hypothesis subtitle file to be scored. They do not contain any explicit sentence boundary information. To calculate traditional MT metrics (BLEU, TER and chrF), we first define reference segments and then align the hypothesis subtitle text to these reference segments by minimizing the edit distance ("Levenshtein alignment") \citep{matusov-etal-2005-evaluating}. Two choices of reference segments are reasonable: 1) subtitle blocks; 2) sentences, split according to simple rules based on sentence-final punctuation, possibly spanning across subtitle blocks. Only for the case of translation from a subtitle template, which preserves subtitle timings, there is a third option, namely to directly use the parallel subtitle blocks as units without any alignment step. This makes the metric sensitive to how translated sentences are distributed among several subtitles, which is a problem a subtitle translation system has to solve.

To evaluate subtitle segmentation quality in isolation, \citet{alvarez2017improving, karakanta-etal-2020-must, karakanta2020point} calculate precision and recall of predicted breaks. Such an analysis is only possible when the subtitle text to be segmented is fixed and the only degree of freedom is the position of breaks. We however consider the general case, where subtitles that differ in text, segmentation and timing are compared and evaluated.

\begin{figure*}[t]
\small
\tt
\fbox{\parbox{\textwidth}{
For the champagne <eol> and brandy you bought me.~<eob> \\
As I recall, the booze put you <eol> to sleep a little prematurely.~<eob> \\
Ladies and gentlemen, <eol> the dance is about to begin.~<eob>
}}
\caption{Example for usage of end-of-line (\texttt{<eol>}) and end-of-block tokens (\texttt{<eob>}) to represent subtitle formatting. Corresponds to right subtitle from Figure \ref{fig:srt_example}. Symbols are adopted from \citet{karakanta-etal-2020-must}.}
\label{fig:breaks_as_tokens_example}
\vspace{-0.3cm}
\end{figure*}

\subsection{Line Break Tokens}
A simple method to extend the baseline metrics to take line and subtitle breaks into account is to insert special tokens at the corresponding positions into the subtitle text \citep{karakanta-etal-2020-42,matusov-etal-2019-customizing}. Figure \ref{fig:breaks_as_tokens_example} shows an example. The automatic metrics treat these tokens as any other word, e.g. BLEU includes them in n-grams, WER and TER count edit operations for them. Therefore, subtitles with a segmentation not matching the reference will get lower scores.

\subsection{Timing-Based Segment Alignment}
\label{subsec:time_alignment}
The time alignment method proposed in \citet{cherry2021subtitle} to calculate t-BLEU is an alternative to Levenshtein hypothesis-to-reference alignment that offers the potential advantage of punishing mistimed words. It uses interpolation of the hypothesis subtitle timings to word-level. Mistimed words may get assigned to a segment without a corresponding reference word, or will even be dropped from the hypothesis if they do not fall into any reference segment.

In this work we consider translation from a template file, thus time alignment is equivalent to using subtitle blocks as unit. However, for the transcription task, where subtitle timings of hypothesis and reference are different, we analyze a variant of WER that operates on "t-BLEU segments", i.e. allows for word matches only if hypothesis and reference word are aligned in time (according to interpolated hypothesis word timings). We refer to this variant as t-WER.

\subsection{New Metric: Subtitle Edit Rate (SubER)}

\begin{figure}[h]
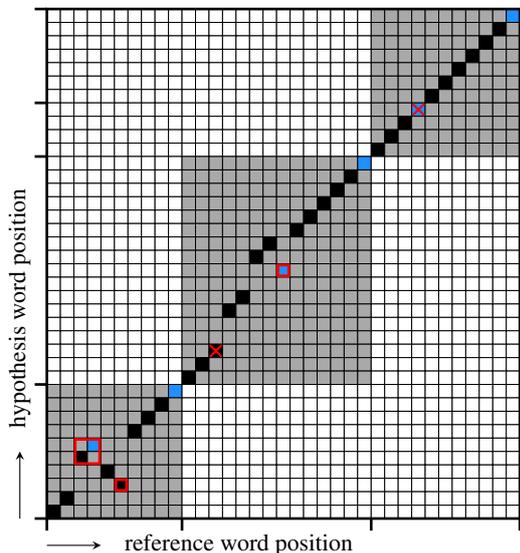

    \centering
    \include{SubER_alignment_example}
    \vspace{-1cm}
    \caption{Visualization of SubER applied to the subtitles from Figure \ref{fig:srt_example} (hypothesis left, reference right). Ticks on the axes indicate subtitle block boundaries. Grey areas show regions of time-overlapping reference and hypothesis subtitles. Word matches, substitutions and shifts are allowed only within those areas. Black squares represent word alignments, blue squares represent break token alignments. Red borders mark shifted phrases, red crosses indicate substitutions. 35 reference words (including breaks), 3 insertions, 2 substitutions, 3 shifts lead to a SubER score of $(3 + 2 + 3) / 35 = 22.86\%$.}
    \label{fig:SubER_alignment_example}
    \vspace{-0.4cm}
\end{figure}

None of the above-mentioned metrics considers \textit{all} of the relevant information present in a subtitle file, namely subtitle text, line segmentation and timing. We therefore propose a new metric called subtitle edit rate (SubER) that attempts to cover all these aspects, and on top avoids segmentation of the subtitle files into aligned hypothesis-reference pairs as a pre-processing step.

We choose TER \citep{snover-etal-2006-study} as the basis of SubER because of its interpretability, especially in the case of post-editing. It corresponds to the number of edit operations, namely substitutions, deletions, insertions and shifts of words that are required to turn the hypothesis text into the reference. Also, it allows for easy integration of segmentation and timing information by extending it with break edit operations and time-alignment constraints.

We define the SubER score to be the minimal possible value of (read "\#" as "number of"):

\begin{equation*}
    \mathrm{SubER} = \frac{\text{\# word edits} + \text{\# break edits} + \text{\# shifts}}{\text{\# reference words} + \text{\# reference breaks}}
\end{equation*}
where
\begin{itemize}
    \itemsep0em
    \item a hypothesis word is only regarded as correct (\textbf{no edit}) if it is part of a subtitle that overlaps in time with the subtitle containing the matching reference word (otherwise edits are required, e.g. deletion + insertion).
    \item \textbf{word edits} are insertions, deletions and substitutions of words, substitutions being only allowed if the hypothesis and reference word are from subtitles that overlap in time. 
    \item \textbf{break edits} are insertions, deletions and substitutions of breaks, treated as additional tokens (\texttt{<eol>} and \texttt{<eob>}) inserted at the positions of the breaks. Substitutions are only allowed between end-of-line and end-of-block, not between a word and a break, and the same time-overlap condition as for word substitution applies.
    \item \textbf{shifts} are movements of one or more adjacent hypothesis tokens to a position of a matching phrase in the reference. Only allowed if all the shifted words come from a hypothesis subtitle that overlaps in time with the subtitle of the matching reference word. The shifted phrase may consist of any combination of words and break tokens.
\end{itemize}

We only consider subtitle timings present in the subtitle files, as opposed to interpolating timings of words as done by \citet{cherry2021subtitle}. This avoids hypothesis words "falling off the edges" of reference subtitles, e.g. in case the hypothesis subtitle starts a fraction of a second early. It also prevents alignment errors originating from the assumption that all words have the same duration.

The time-overlap condition can be thought of as constraining the search space for Levenshtein-distance calculation. Figure \ref{fig:SubER_alignment_example} visualizes this for the subtitles from Figure \ref{fig:srt_example}. In the white areas no word matches are allowed, this can be exploited for an efficient implementation. The last two hypothesis subtitles overlap with the last reference subtitle and therefore form a single time-aligned region. The shifted 2-word phrase in the bottom left region is "\texttt{champagne <eol>}", showcasing that words and breaks can be shifted in a single operation. In the center region we see the substitution of "\texttt{recall}" with "\texttt{remember}", the inserted (i.e. unaligned) hypothesis words "\texttt{it}", "\texttt{was}" and "\texttt{that}", and a shift of the line break to a different position. The break substitution in the upper right region corresponds to the fact that the last block of the right  subtitles in Figure \ref{fig:srt_example} is split into two, i.e. end-of-line is replaced by end-of-block.

\subsubsection{Implementation Details}

We modify the TER implementation of SacreBLEU \citep{post-2018-call} to implement SubER. We adopt the approximation of greedily searching for the best shift until no further reduction of the edit distance can be achieved \citep{snover-etal-2006-study}. Break tokens (\texttt{<eol>} and \texttt{<eob>}) are inserted into the input text. String comparisons between hypothesis and reference words are replaced by a function additionally checking the time-overlap condition. To make SubER calculation feasible for large subtitle files we split hypothesis and reference into parts at time positions where both agree that no subtitle is displayed. The number of edit operations is then added up for all parts. By definition this does not affect the metric score, in contrast to e.g. segmenting into sentence vs. subtitle blocks when calculating BLEU (Section \ref{subsec:base_metrics}).

\section{Human Evaluation}
To analyze the expressiveness of SubER we conduct a human post-editing experiment on both subtitles automatically generated from audio, as well as automatic translations of subtitle text files. For each of the two post-editing tasks we employ three professional subtitlers with multiple years of experience in the subtitling industry. We evaluate how well automatic metric scores correlate with their post-editing effort and their MT quality judgements.

There exists previous work measuring the productivity gains from post-editing automatic subtitles under the aspect of MT quality \citep{etchegoyhen-etal-2014-machine,bywood2017embracing,koponen-etal-2020-mt} and segmentation quality \citep{alvarez-etal-2016-impact,alvarez2017improving,matusov-etal-2019-customizing}, but to the best of our knowledge we conduct the first study with the goal of evaluating an automatic quality metric for subtitling.


\subsection{Data}
We perform our experiment using one episode from each of the following shows:
\begin{itemize}
    \itemsep0em
    \item \textit{Master of None:} a comedy-drama series
    \item \textit{Midnight Mass:} a supernatural horror series 
    \item \textit{Peaky Blinders:} an early 20th century British gangster drama 
\end{itemize}

Each of the three videos has a duration of approximately 55 minutes. They are originally in English, for translation we choose Spanish as the target language.
We use pre-existing English subtitles as template files for human translation, and also as the reference when scoring automatic transcriptions. Pre-existing Spanish subtitles, which follow the English template, are used as reference for MT output.

To gather data points for which we can compare post-editing effort with automatic scores, we manually split the videos into segments of roughly 1 minute, each containing 15 subtitle blocks and 103 words on average. We keep the first 15 minutes of each video as one large segment where we measure baseline speed of the subtitlers. Excluding these, we end up with 35, 38 and 37 segments for the videos, respectively, amounting to a total of 110 source-target reference subtitle pairs.

\subsection{Automatic Subtitling Systems}

For human post-editing, we create automatic English and Spanish subtitle files. We use several different subtitling systems to obtain evaluation data with a wider variety. The systems differ in ASR/MT, punctuation and segmentation quality.

We create a single automatic English and Spanish subtitle file for each video, each containing segments coming from different automatic subtitling systems. The subtitlers did not know about any of the details on how these files were created to avoid any bias.

\subsubsection{Transcription Systems}\label{sec:transcr_systems}

To create automatic English subtitles from the audio track of the video we use three different systems:
\begin{enumerate}
    \itemsep0em
    \item A hybrid ASR system, the output of which is punctuated and cased by a bi-directional LSTM model and then split into lines and subtitles using a beam search decoder that combines scores of a neural segmentation model and hard subtitling constraints, based on the algorithm proposed by \citet{matusov-etal-2019-customizing};
    \item same as 1., but without using a neural model for subtitle segmentation;
    \item an online provider offering automatic transcription in SRT format.
\end{enumerate}
We transcribe an equal number of video segments with each of the three systems and combine them into a single subtitle file which is delivered to the subtitlers for post-editing. The first segment of 15 minutes is not transcribed automatically. Instead, the subtitlers are asked to transcribe it from scratch to measure their baseline productivity. 

\subsubsection{Translation Systems}

To create Spanish subtitles we translate the pre-existing English subtitles with 5 different systems:
\begin{enumerate}
    \itemsep0em
    \item A Transformer-based MT system, the output of which is split into lines and subtitles using a neural segmentation model and hard subtitling constraints;
    \item same as 1., but without using a neural model for subtitle segmentation;
    \item same as 1., but with additional inputs for length control and genre, similarly to the systems proposed in~\cite{schioppa-etal-2021-controlling, matusov-etal-2020-flexible}; 
    \item an LSTM-based MT system with lower quality than 1., but also using the neural segmentation model;
    \item an online provider offering subtitle translation in SRT format.
\end{enumerate}
Also here, we distribute the video segments among the systems such that each system contributes a roughly equal portion of the assembled MT subtitle file delivered to the translators. We extract full sentences from the source subtitle file based on punctuation before translation. The first 15 minute segment of each video is translated directly from the source template without access to MT output to measure baseline productivity of the translators.

\subsection{Methodology}

\subsubsection{Productivity Gain Measurement}

For both transcription and translation, we ask the subtitlers to measure the time $t_n$ (in minutes) spent to post-edit each of the 110 video segments. As a measure of post-editing productivity $P_n$ we compute the number of subtitles $S_n$ created per minute of work for the $n$-th segment:
\begin{equation}
    P_n = \frac{S_n}{t_n}
\end{equation}
To make these values comparable between subtitlers we normalize them using the subtitler's baseline speed $P_\mathrm{base}$. It is computed by averaging the productivity in the first 15-minute segment $P_1$, where the subtitlers work from scratch, over all three videos.
Finally, we average the normalized productivities across the three subtitlers $h=1,2,3$ per task to get an average post-editing productivity gain for segment $n$:
\begin{equation}
    \hat{P}_n = \frac{1}{3}\sum_{h=1}^3{\frac{P_{n,h}}{P_{\mathrm{base},h}}}
\end{equation}
To evaluate the expressiveness of a given metric we compute the
Spearman's rank correlation coefficient $r_s$ between the per-segment metric scores and $\hat{P}_n$
for all segments of all three videos. We choose Spearman's correlation in favour of Pearson's correlation because subtitle quality varies a lot for different video segments and different systems, and we don't expect the metrics to behave linearly in this range.

\subsubsection{Direct Assessment}
\label{subsubsec:direct_assessment}

For the translation task we additionally gather direct assessment scores for each segment. For this we ask the translators to give two scores (referred to as $U_n$ and $Q_n$, respectively) according to the following descriptions:

\begin{enumerate}
    \itemsep0em
    \item "Rate the overall \textbf{usefulness} of the automatically translated subtitles in this segment for post-editing purposes on a scale from 0 (completely useless) to 100 (perfect, not a single change needed)."
    \item "Rate the overall \textbf{quality} of the automatically translated subtitles in this segment as perceived \textit{by a viewer} on a scale from 0 (completely incomprehensible) to 100 (perfect, completely fluent and accurate). The score should reflect how well the automatic translation conveys the semantics of the original subtitles, and should also reflect how well the translated subtitles are formatted."
\end{enumerate}

These scores are standardized into $z$-scores by subtracting the average and dividing by the standard deviation of scores per translator. Finally, we average the $z$-scores across the three translators to get expected usefulness and quality assessment scores for each segment, which we will refer to as $\hat{U}_n$ and $\hat{Q}_n$, respectively. 

\subsection{Results}

\subsubsection{Post-Editing of English Transcription}

\begin{table*}[h!]
\centering
\scalebox{0.9}{
\begin{tabular}{| l || c | c | c || c |} 
 \hline
 \textbf{Metric} & \textbf{Subtitler A} & \textbf{Subtitler B} & \textbf{Subtitler C} &  \textbf{Combined} \\
 \hline
 \hline
 WER & -0.731 & -0.494 & -0.499 & -0.676 \\ 
 + case/punct & -0.671 & \textbf{-0.512} & -0.509 & -0.650 \\
 + break tokens & -0.725 & -0.494 & -0.512 & -0.678 \\
 t-WER & -0.661 & -0.440 & -0.476 & -0.625 \\
 TER-br & -0.573 & -0.489 & -0.434 & -0.562 \\
 \hline
 SubER (ours) & \textbf{-0.746} & -0.506 & \textbf{-0.517} & \textbf{-0.692} \\
 + case/punct & -0.670 & -0.507 & -0.500 & -0.645 \\
 - break tokens & -0.741 & -0.495 & -0.502 & -0.682 \\
 
 \hline
\end{tabular}}
\caption{Spearman's correlation $r_s$ between automatic metric scores and post-editing productivity gains $P_n$ on all 110 video segments for the \textbf{English transcription task}. The last column shows correlation to the productivity gain averaged across subtitlers $\hat{P}_n$.}
\vspace{-0.3cm}
\label{table:asr_correlations}
\end{table*}
The baseline productivities $P_\mathrm{base}$ of the three subtitlers A, B and C when transcribing the first 15 minutes of each video from scratch are 3.4, 2.8 and 2.7 subtitles per minute of work, respectively. Post-editing changes their productivities to 3.9, 2.6 and 3.1 subtitles per minute on average for the other segments, meaning subtitlers A and C work faster when post-editing automatic subtitles, while subtitler B does not benefit from them.

Table \ref{table:asr_correlations} shows the analysis of  the correlation between automatic metric scores and productivity gains, calculated for each of the 110 one-minute video segments.
Word error rate (WER) can predict the averaged productivity gain $\hat{P}_n$ with a Spearman's correlation of $-0.676$. This confirms the natural assumption that the more words the ASR system recognized correctly in a given segment, the less time is necessary for post-editing. Subtitler A's post-editing gains are more predictable than those of the other two subtitlers. This indicates that the subtitlers have different workflows and do not make use of the automatic subtitles with the same consistency.

Row 2 shows that making WER case-sensitive and keeping punctuation marks as part of the words does not improve correlation consistently. Although we believe that casing and punctuation errors harm subtitle quality, these errors might not have a significant impact on post-editing time because correcting them requires changing single characters only. Row 3 shows that extending the original WER definition by simply inserting end-of-line and end-of-block tokens into the text does not lead to improvements either. This can be explained by the fact that the original WER algorithm allows for substitution of break symbols with words. Such substitutions have no meaningful interpretation. Also, it does not support shifts of break symbols, which leads to breaks at wrong positions being punished more than completely missing ones.

Our proposed metric SubER achieves the overall best correlation of $-0.692$.
We attribute this in part to a proper way of handling segmentation information: without it, as shown in the last row of Table~\ref{table:asr_correlations}, the correlation is lower. Unfortunately, for the same reasons as for the case of WER, we have to apply SubER to lower-cased text - as it is the default setting for the TER metric - to avoid a drop in correlation.

Correlations for t-WER (see Section \ref{subsec:time_alignment}) suggest that a word-level time-alignment 
using interpolation 
may result in misalignments which are punished too harsh in comparison to which mistimings are still tolerated by the post-editors. This supports our design choice of using subtitle-level timings for SubER.

Finally, we include TER-br from \citet{karakanta-etal-2020-42} in the results. It is a variant of TER + break tokens where each real word is replaced by a mask token. Given that the metric has no access to the actual words it achieves surprisingly high correlations. This shows that the subtitle formatting defined by the number of subtitle blocks, number of lines and number of words per line is in itself an important feature affecting the post-editing effort.

\subsubsection{Post-Editing of Spanish Translation}
\begin{table*}[h!]
\small
\centering
\begin{tabular}{| l || c c c | c c c | c c c || c c c |} 
 \hline
 \textbf{Metric} & \multicolumn{3}{|c|}{\textbf{Subtitler D}} & \multicolumn{3}{|c|}{\textbf{Subtitler E}} & \multicolumn{3}{|c||}{\textbf{Subtitler F}} & \multicolumn{3}{|c|}{\textbf{Combined}} \\
 & $P_n$ & $U_n$ & $Q_n$ & $P_n$ & $U_n$ & $Q_n$ & $P_n$ & $U_n$ & $Q_n$ & $\hat{P}_n$ & $\hat{U}_n$ & $\hat{Q}_n$ \\
 \hline
 \hline
 \textbf{Subtitle-level} & & & & & & & & & & & & \\
  BLEU & ~0.03 & ~0.34 & ~0.52 &  ~0.22 & ~0.21 & ~0.39 &  ~0.07 & ~0.58 & ~0.49 & ~0.172 & ~0.541 & ~0.595 \\ 
 + break tokens & ~0.04 & ~0.35 & ~0.53 &  ~0.22 & ~0.24 & ~0.43 &  ~0.12 & ~0.58 & ~0.46 & ~0.210 & ~0.554 & ~0.595 \\
 TER & ~0.03 & -0.35 & -0.54 &  -0.22 & -0.23 & -0.41 &  -0.11 & -0.63 & -0.51 & -0.182 & -0.554 & -0.618 \\
 + break tokens & ~0.00 & -0.36 & -0.54 &  -0.23 & -0.24 & -0.41 &  -0.10 & -0.61 & -0.50 & -0.200 & -0.558 & -0.606 \\
 \hline
 \textbf{Sentence-level} & & & & & & & & & & & & \\
 BLEU & -0.03 & ~0.31 & ~0.51 &  ~0.21 & ~0.13 & ~0.33 &  ~0.04 & ~0.60 & ~0.51 & ~0.126 & ~0.494 & ~0.573 \\ 
 + break tokens & ~0.02 & ~0.35 & ~0.55 &  ~0.25 & ~0.22 & ~0.43 &  ~0.16 & ~0.63 & \textbf{~0.55} & ~0.240 & ~0.583 & \textbf{~0.659} \\
 TER & ~0.07 & -0.32 & -0.52 &  -0.22 & -0.14 & -0.34 &  -0.07 & -0.59 & -0.48 & -0.133 & -0.484 & -0.559 \\
 + break tokens & ~0.00 & -0.36 & -0.55 &  -0.25 & -0.19 & -0.38 &  -0.13 & -0.58 & -0.45 & -0.218 & -0.515 & -0.574 \\ 
 chrF & -0.09 & ~0.26 & ~0.52 &  ~0.21 & ~0.10 & ~0.28 &  ~0.04 & ~0.64 & ~0.51 & ~0.104 & ~0.483 & ~0.556 \\
 TER-br & ~0.03 & -0.32 & -0.42 &  -0.11 & -0.07 & -0.24 &  -0.13 & -0.43 & -0.40 & -0.137 & -0.345 & -0.426 \\
 \hline
 SubER (ours) & -0.06 & \textbf{-0.38} & \textbf{-0.57} & \textbf{-0.27} & \textbf{-0.28} & \textbf{-0.47} & -0.16 & -0.61 & -0.52 & \textbf{-0.274} & \textbf{-0.591} & -0.651 \\
 + case/punct & ~0.00 & -0.36 & -0.56 &  -0.25 & -0.23 & -0.42 &  -0.15 & -0.61 & -0.49 & -0.237 & -0.554 & -0.612 \\
 - break tokens & ~0.02 & -0.34 & -0.54 & -0.24 & -0.25 & -0.44 & -0.11 & \textbf{-0.65} & \textbf{-0.55} & -0.197 & -0.572 & -0.645 \\

 \hline
\end{tabular}
\caption{Spearman's correlation $r_s$ between automatic metric scores and $P_n$, $U_n$ and $Q_n$ on all 110 video segments for the \textbf{English$\rightarrow$Spanish translation task}. $P_n$ are segment-wise productivity gains from post-editing measured in subtitles per minute of work. $U_n$ and $Q_n$ are segment-wise usefulness and quality scores, respectively, which the subtitlers assigned to the automatically generated subtitle segments.}
\vspace{-0.3cm}
\label{table:mt_correlations}
\end{table*}

Baseline productivities $P_\mathrm{base}$ of the translators D, E and F are 1.9, 1.8 and 1.1 subtitles per minute, respectively. On average, their productivity changes to 1.6, 2.0 and 1.1 when post-editing, meaning only subtitler B gains consistently. Subtitler A is more productive on one of the videos, but slows down significantly for the other two.

Table \ref{table:mt_correlations} shows performances of the different MT metrics. In addition to post-edit effort, we show how well the metrics agree with human judgments of the usefulness and quality (see Section \ref{subsubsec:direct_assessment}) for each of the 110 one-minute video segments.

Overall, the correlation of productivity gains is much lower than for the transcription task. This can be explained by the fact that a translator has more freedom
than a transcriber. 
The translator's word choices are influenced by clues outside the scope of the translated text,
like the style of language and references to other parts of the plot. Sometimes even research is required (e.g. bible verses for \textit{Midnight Mass}).
Despite this, the subjectively perceived usefulness $\hat{U}_n$ of the automatic subtitles for post-editing can be predicted from automatic scores with a Spearman's correlation of up to $-0.591$. The quality judgement $\hat{Q}_n$ shows even higher correlations of up to $0.659$. 

We compare the baseline MT metrics BLEU and TER when applied to the subtitle block-level vs. the sentence-level. We note that BLEU on subtitle-level is identical to t-BLEU \cite{cherry2021subtitle} for the considered case of template translation, where timestamps in hypothesis and reference are identical. Overall, BLEU and TER perform similarly. For both, evaluation on subtitle-level outperforms evaluation on sentence-level. This is because the sentence-pairs extracted from the subtitle files preserve no formatting information, while using subtitle blocks as units is sensitive to how words of a sentence are distributed among subtitles after translation, especially in case of word re-ordering.

Extending BLEU and TER with break tokens to take subtitle segmentation into account shows only minor improvements for the subtitle-level, but significantly improves correlations for the sentence-level.
This could be attributed to the extended context after end-of-block tokens that is not available for scoring on subtitle-level. Especially the way "BLEU + break tokens" punishes n-grams that are disrupted by an erroneous line break seems to lead to good results.


Our proposed metric SubER consistently outperforms all considered baseline metrics except for sentence-level BLEU with break tokens, which has a higher correlation for $\hat{Q}_n$ and for the scores given by subtitler F. For this subtitler we also observe that calculating SubER without break tokens improves results.
In fact, subtitler F stated that moving around text is not a taxing procedure for him as he is very proficient with keyboard commands. For the other subtitlers, break tokens as part of the metric are shown to have a clear positive effect.

\subsubsection{System-level Results}
For both transcription and translation we have a pair of systems which differ only in subtitle segmentation (systems 1 and 2). We expect the system using a neural segmentation model to perform better overall. By definition, WER cannot distinguish between the transcription systems, scores for both are 40.6, 14.2 and 29.5 (\%) for the three videos \textit{Master of None}, \textit{Midnight Mass} and \textit{Peaky Blinders}, respectively. (High WER on \textit{Master of None} is caused by colloquial and mumbling speech.) SubER scores for system 1 are 46.4, 20.3 and 33.1, for system 2 they are 47.3, 22.1 and 34.7. This means, for all videos SubER scores are able to reflect the better segmentation quality of system 1.

The same is true for translation: sentence-level BLEU scores are the same for systems 1 and 2, namely 18.9, 26.7 and 37.9 for the three videos. SubER scores for the system with neural segmentation are 65.1, 56.5 and 41.8, whereas the system without it gets worse scores of 67.4, 60.5 and 46.9.

\section{Release of Code}
We release the code to calculate the SubER metric as part of an open-source subtitle evaluation toolkit\footnote{\url{https://github.com/apptek/SubER}} to encourage its use in the research community as well as the media industry and to further promote research of automatic subtitling systems.

In addition to SubER, the toolkit implements all baseline metrics used in Table \ref{table:asr_correlations} and \ref{table:mt_correlations}, as well as t-BLEU \citep{cherry2021subtitle}. This includes implementations of hypothesis to reference alignment via the Levenshtein algorithm (Section \ref{subsec:base_metrics}) or via interpolated word timings (Section \ref{subsec:time_alignment}).
We use the JiWER\footnote{\url{https://github.com/jitsi/jiwer}} Python package for word error rate calculations and SacreBLEU \citep{post-2018-call} to compute BLEU, TER and chrF values.

All metrics can be calculated directly from SRT input files. Support for other subtitle file formats will be added on demand.

\section{Conclusion}
In this work, we proposed SubER -- a novel metric for evaluating quality of automatically generated intralingual and interlingual subtitles. The metric is based on edit distance with shifts, but considers not only the automatically transcribed or translated text, but also subtitle timing and line segmentation information. It can be used to compare an automatically generated subtitle file to a human-generated one even if the two files contain a different number of subtitles with different timings.

A thorough evaluation by professional subtitlers confirmed that SubER correlates well with their transcription post-editing effort and direct assessment scores of translations. In most cases, SubER shows highest correlation as compared to metrics that evaluate either the quality of the text alone, or use different approaches to integrate subtitle timing and segmentation information.

The source code for SubER will be publicly released for the benefit of speech recognition and speech translation research communities, as well as the media and entertainment industry.


\bibliography{anthology,custom}

\begin{thebibliography}{22}
\expandafter\ifx\csname natexlab\endcsname\relax\def\natexlab#1{#1}\fi

\bibitem[{{\'A}lvarez et~al.(2016){\'A}lvarez, Balenciaga, del Pozo, Arzelus,
  Matamala, and Mart{\'\i}nez-Hinarejos}]{alvarez-etal-2016-impact}
Aitor {\'A}lvarez, Marina Balenciaga, Arantza del Pozo, Haritz Arzelus, Anna
  Matamala, and Carlos-D. Mart{\'\i}nez-Hinarejos. 2016.
\newblock \href {https://aclanthology.org/L16-1487} {Impact of automatic
  segmentation on the quality, productivity and self-reported post-editing
  effort of intralingual subtitles}.
\newblock In \emph{Proceedings of the Tenth International Conference on
  Language Resources and Evaluation ({LREC}'16)}, pages 3049--3053,
  Portoro{\v{z}}, Slovenia. European Language Resources Association (ELRA).

\bibitem[{Alvarez et~al.(2017)Alvarez, Mart{\'\i}nez-Hinarejos, Arzelus,
  Balenciaga, and del Pozo}]{alvarez2017improving}
Aitor Alvarez, Carlos-D Mart{\'\i}nez-Hinarejos, Haritz Arzelus, Marina
  Balenciaga, and Arantza del Pozo. 2017.
\newblock \href
  {https://riunet.upv.es/bitstream/handle/10251/104008/Improving_the_Automatic_Segmentation_of_Subtitles.pdf}
  {Improving the automatic segmentation of subtitles through conditional random
  field}.
\newblock \emph{Speech Communication}, 88:83--95.

\bibitem[{Bywood et~al.(2017)Bywood, Georgakopoulou, and
  Etchegoyhen}]{bywood2017embracing}
Lindsay Bywood, Panayota Georgakopoulou, and Thierry Etchegoyhen. 2017.
\newblock \href
  {https://westminsterresearch.westminster.ac.uk/download/9a38cde4c5b97fdc82bb0f267bdd136e7a7610d2a08b9e25bf6c952e71f71399/191070/Bywood\%20et\%20al.\%20OK.pdf}
  {Embracing the threat: machine translation as a solution for subtitling}.
\newblock \emph{Perspectives}, 25(3):492--508.

\bibitem[{Cherry et~al.(2021)Cherry, Arivazhagan, Padfield, and
  Krikun}]{cherry2021subtitle}
Colin Cherry, Naveen Arivazhagan, Dirk Padfield, and Maxim Krikun. 2021.
\newblock \href
  {https://www.isca-speech.org/archive/pdfs/interspeech_2021/cherry21_interspeech.pdf}
  {Subtitle translation as markup translation}.
\newblock \emph{Proc. Interspeech 2021}, pages 2237--2241.

\bibitem[{Etchegoyhen et~al.(2014)Etchegoyhen, Bywood, Fishel, Georgakopoulou,
  Jiang, van Loenhout, del Pozo, Mau{\v{c}}ec, Turner, and
  Volk}]{etchegoyhen-etal-2014-machine}
Thierry Etchegoyhen, Lindsay Bywood, Mark Fishel, Panayota Georgakopoulou, Jie
  Jiang, Gerard van Loenhout, Arantza del Pozo, Mirjam~Sepesy Mau{\v{c}}ec,
  Anja Turner, and Martin Volk. 2014.
\newblock \href
  {http://www.lrec-conf.org/proceedings/lrec2014/pdf/463_Paper.pdf} {Machine
  translation for subtitling: A large-scale evaluation}.
\newblock In \emph{Proceedings of the Ninth International Conference on
  Language Resources and Evaluation ({LREC}'14)}, pages 46--53, Reykjavik,
  Iceland. European Language Resources Association (ELRA).

\bibitem[{Karakanta et~al.(2020{\natexlab{a}})Karakanta, Negri, and
  Turchi}]{karakanta-etal-2020-42}
Alina Karakanta, Matteo Negri, and Marco Turchi. 2020{\natexlab{a}}.
\newblock \href {https://doi.org/10.18653/v1/2020.iwslt-1.26} {Is 42 the answer
  to everything in subtitling-oriented speech translation?}
\newblock In \emph{Proceedings of the 17th International Conference on Spoken
  Language Translation}, pages 209--219, Online. Association for Computational
  Linguistics.

\bibitem[{Karakanta et~al.(2020{\natexlab{b}})Karakanta, Negri, and
  Turchi}]{karakanta-etal-2020-must}
Alina Karakanta, Matteo Negri, and Marco Turchi. 2020{\natexlab{b}}.
\newblock \href {https://aclanthology.org/2020.lrec-1.460} {{M}u{ST}-cinema: a
  speech-to-subtitles corpus}.
\newblock In \emph{Proceedings of the 12th Language Resources and Evaluation
  Conference}, pages 3727--3734, Marseille, France. European Language Resources
  Association.

\bibitem[{Karakanta et~al.(2020{\natexlab{c}})Karakanta, Negri, and
  Turchi}]{karakanta2020point}
Alina Karakanta, Matteo Negri, and Marco Turchi. 2020{\natexlab{c}}.
\newblock \href {http://ceur-ws.org/Vol-2769/paper_78.pdf} {Point break:
  Surfing heterogeneous data for subtitle segmentation}.
\newblock In \emph{CLiC-it}.

\bibitem[{Koponen et~al.(2020)Koponen, Sulubacak, Vitikainen, and
  Tiedemann}]{koponen-etal-2020-mt}
Maarit Koponen, Umut Sulubacak, Kaisa Vitikainen, and J{\"o}rg Tiedemann. 2020.
\newblock \href {https://aclanthology.org/2020.eamt-1.13} {{MT} for subtitling:
  User evaluation of post-editing productivity}.
\newblock In \emph{Proceedings of the 22nd Annual Conference of the European
  Association for Machine Translation}, pages 115--124, Lisboa, Portugal.
  European Association for Machine Translation.

\bibitem[{Matusov et~al.(2005)Matusov, Leusch, Bender, and
  Ney}]{matusov-etal-2005-evaluating}
Evgeny Matusov, Gregor Leusch, Oliver Bender, and Hermann Ney. 2005.
\newblock \href {https://aclanthology.org/2005.iwslt-1.19} {Evaluating machine
  translation output with automatic sentence segmentation}.
\newblock In \emph{Proceedings of the Second International Workshop on Spoken
  Language Translation}, Pittsburgh, Pennsylvania, USA.

\bibitem[{Matusov et~al.(2019)Matusov, Wilken, and
  Georgakopoulou}]{matusov-etal-2019-customizing}
Evgeny Matusov, Patrick Wilken, and Yota Georgakopoulou. 2019.
\newblock \href {https://doi.org/10.18653/v1/W19-5209} {Customizing neural
  machine translation for subtitling}.
\newblock In \emph{Proceedings of the Fourth Conference on Machine Translation
  (Volume 1: Research Papers)}, pages 82--93, Florence, Italy. Association for
  Computational Linguistics.

\bibitem[{Matusov et~al.(2020)Matusov, Wilken, and
  Herold}]{matusov-etal-2020-flexible}
Evgeny Matusov, Patrick Wilken, and Christian Herold. 2020.
\newblock \href {https://aclanthology.org/2020.amta-user.10} {Flexible
  customization of a single neural machine translation system with
  multi-dimensional metadata inputs}.
\newblock In \emph{Proceedings of the 14th Conference of the Association for
  Machine Translation in the Americas (Volume 2: User Track)}, pages 204--216,
  Virtual. Association for Machine Translation in the Americas.

\bibitem[{Papineni et~al.(2002)Papineni, Roukos, Ward, and
  Zhu}]{papineni-etal-2002-bleu}
Kishore Papineni, Salim Roukos, Todd Ward, and Wei-Jing Zhu. 2002.
\newblock \href {https://doi.org/10.3115/1073083.1073135} {{B}leu: a method for
  automatic evaluation of machine translation}.
\newblock In \emph{Proceedings of the 40th Annual Meeting of the Association
  for Computational Linguistics}, pages 311--318, Philadelphia, Pennsylvania,
  USA. Association for Computational Linguistics.

\bibitem[{Pedersen(2017)}]{pedersen2017}
J.~Pedersen. 2017.
\newblock \href {https://www.jostrans.org/issue28/art_pedersen.pdf} {The {FAR}
  model: assessing quality in interlingual subtitling}.
\newblock In \emph{Journal of Specialized Translation}, volume~18, pages
  210--229.

\bibitem[{Popovi{\'c}(2015)}]{popovic-2015-chrf}
Maja Popovi{\'c}. 2015.
\newblock \href {https://doi.org/10.18653/v1/W15-3049} {chr{F}: character
  n-gram {F}-score for automatic {MT} evaluation}.
\newblock In \emph{Proceedings of the Tenth Workshop on Statistical Machine
  Translation}, pages 392--395, Lisbon, Portugal. Association for Computational
  Linguistics.

\bibitem[{Post(2018)}]{post-2018-call}
Matt Post. 2018.
\newblock \href {https://doi.org/10.18653/v1/W18-6319} {A call for clarity in
  reporting {BLEU} scores}.
\newblock In \emph{Proceedings of the Third Conference on Machine Translation:
  Research Papers}, pages 186--191, Brussels, Belgium. Association for
  Computational Linguistics.

\bibitem[{Rei et~al.(2020)Rei, Stewart, Farinha, and
  Lavie}]{rei-etal-2020-comet}
Ricardo Rei, Craig Stewart, Ana~C Farinha, and Alon Lavie. 2020.
\newblock \href {https://doi.org/10.18653/v1/2020.emnlp-main.213} {{COMET}: A
  neural framework for {MT} evaluation}.
\newblock In \emph{Proceedings of the 2020 Conference on Empirical Methods in
  Natural Language Processing (EMNLP)}, pages 2685--2702, Online. Association
  for Computational Linguistics.

\bibitem[{Romero-Fresco and P\"ochhacker(2017)}]{romero2017}
P.~Romero-Fresco and F.~P\"ochhacker. 2017.
\newblock \href
  {https://lans-tts.uantwerpen.be/index.php/LANS-TTS/article/view/438} {Quality
  assessment in interlingual live subtitling: The {NTR} model.}
\newblock In \emph{Linguistica Antverpiensia, New Series: Themes in Translation
  Studies}, volume~16, pages 149--167.

\bibitem[{Romero-Fresco and Pérez(2015)}]{romeroperez2015}
P.~Romero-Fresco and J.M. Pérez. 2015.
\newblock \href {https://doi.org/10.1057/9781137552891_3} {Accuracy rate in
  live subtitling: The {NER} model}.
\newblock In \emph{Audiovisual Translation in a Global Context. Palgrave
  Studies in Translating and Interpreting}. R.B., Cintas J.D. (eds), Palgrave
  Macmillan, London.

\bibitem[{Schioppa et~al.(2021)Schioppa, Vilar, Sokolov, and
  Filippova}]{schioppa-etal-2021-controlling}
Andrea Schioppa, David Vilar, Artem Sokolov, and Katja Filippova. 2021.
\newblock \href {https://doi.org/10.18653/v1/2021.emnlp-main.535} {Controlling
  machine translation for multiple attributes with additive interventions}.
\newblock In \emph{Proceedings of the 2021 Conference on Empirical Methods in
  Natural Language Processing}, pages 6676--6696, Online and Punta Cana,
  Dominican Republic. Association for Computational Linguistics.

\bibitem[{Snover et~al.(2006)Snover, Dorr, Schwartz, Micciulla, and
  Makhoul}]{snover-etal-2006-study}
Matthew Snover, Bonnie Dorr, Rich Schwartz, Linnea Micciulla, and John Makhoul.
  2006.
\newblock \href {https://aclanthology.org/2006.amta-papers.25} {A study of
  translation edit rate with targeted human annotation}.
\newblock In \emph{Proceedings of the 7th Conference of the Association for
  Machine Translation in the Americas: Technical Papers}, pages 223--231,
  Cambridge, Massachusetts, USA. Association for Machine Translation in the
  Americas.

\bibitem[{Zhang et~al.(2019)Zhang, Kishore, Wu, Weinberger, and
  Artzi}]{zhang2019bertscore}
Tianyi Zhang, Varsha Kishore, Felix Wu, Kilian~Q Weinberger, and Yoav Artzi.
  2019.
\newblock \href {https://openreview.net/pdf?id=SkeHuCVFDr} {{BERTScore}:
  Evaluating text generation with {BERT}}.
\newblock In \emph{International Conference on Learning Representations}.

\end{thebibliography}
\bibliographystyle{acl_natbib}

\appendix


\end{document}